\documentclass[11pt,a4paper]{article}
\usepackage[hyperref]{acl2019}
\usepackage{times}
\usepackage{latexsym}
\usepackage{graphicx}  

\usepackage{url}
\usepackage{amsfonts}
\usepackage{amsmath}
\usepackage{multirow}
\usepackage{tabularx}
\usepackage{pgfplots} 
\usepackage{booktabs} 

\newcolumntype{C}[1]{>{\centering\arraybackslash}p{#1}}

\aclfinalcopy 
\def\aclpaperid{2888} 

\newcommand{\vpara}[1]{\vspace{0.1in}\noindent\textbf{#1 }}
\newcommand{\para}[1]{\vspace{0.01in}\noindent\textbf{#1 }}

\title{Multi-Channel Graph Neural Network for Entity Alignment}

\author{Yixin Cao$^1$ \quad Zhiyuan Liu$^2$ \quad Chengjiang Li$^3$\\
\textbf{Zhiyuan Liu$^3$ \quad Juanzi Li$^3$ \quad Tat-Seng Chua$^1$}\\
$^1$School of Computing, National University of Singapore, Singapore\\
$^2$School of Science, Xi'an Jiaotong University, Xi'an, China\\
$^3$Department of CST, Tsinghua University, Beijing, China\\
{\tt \{caoyixin2011,acharkq,iamlockelightning\}@gmail.com} \\
{\tt \{liuzy,lijuanzi\}@tsinghua.edu.cn, dcscts@nus.edu.sg} \\
}

\date{}

\begin{document}
\maketitle
\begin{abstract}
  Entity alignment typically suffers from the issues of structural heterogeneity and limited seed alignments. In this paper, we propose a novel Multi-channel Graph Neural Network model (MuGNN) to learn alignment-oriented knowledge graph (KG) embeddings by robustly encoding two KGs via multiple channels. Each channel encodes KGs via different relation weighting schemes with respect to self-attention towards KG completion and cross-KG attention for pruning exclusive entities respectively, which are further combined via pooling techniques. Moreover, we also infer and transfer rule knowledge for completing two KGs consistently. MuGNN is expected to reconcile the structural differences of two KGs, and thus make better use of seed alignments. Extensive experiments on five publicly available datasets demonstrate our superior performance (5\% Hits@1 up on average). Source code and data used in the experiments can be accessed at \url{https://github.com/thunlp/MuGNN}.
\end{abstract}

\section{Introduction}
\label{sec:intro}
Knowledge Graphs (KGs) store the world knowledge in the form of directed graphs, where nodes denote entities and edges are their relations. Since it was proposed, many KGs are constructed (e.g., YAGO~\cite{rebele2016yago}) to provide structural knowledge for different applications and languages. These KGs usually contain complementary contents, attracting researchers to integrate them into a unified KG, which shall benefit many knowledge driven tasks, such as information extraction~\cite{cao2018neural} and recommendation~\cite{wang2018explainable}.

\begin{figure}[htb]
	\centerline{\includegraphics[width=0.45\textwidth]{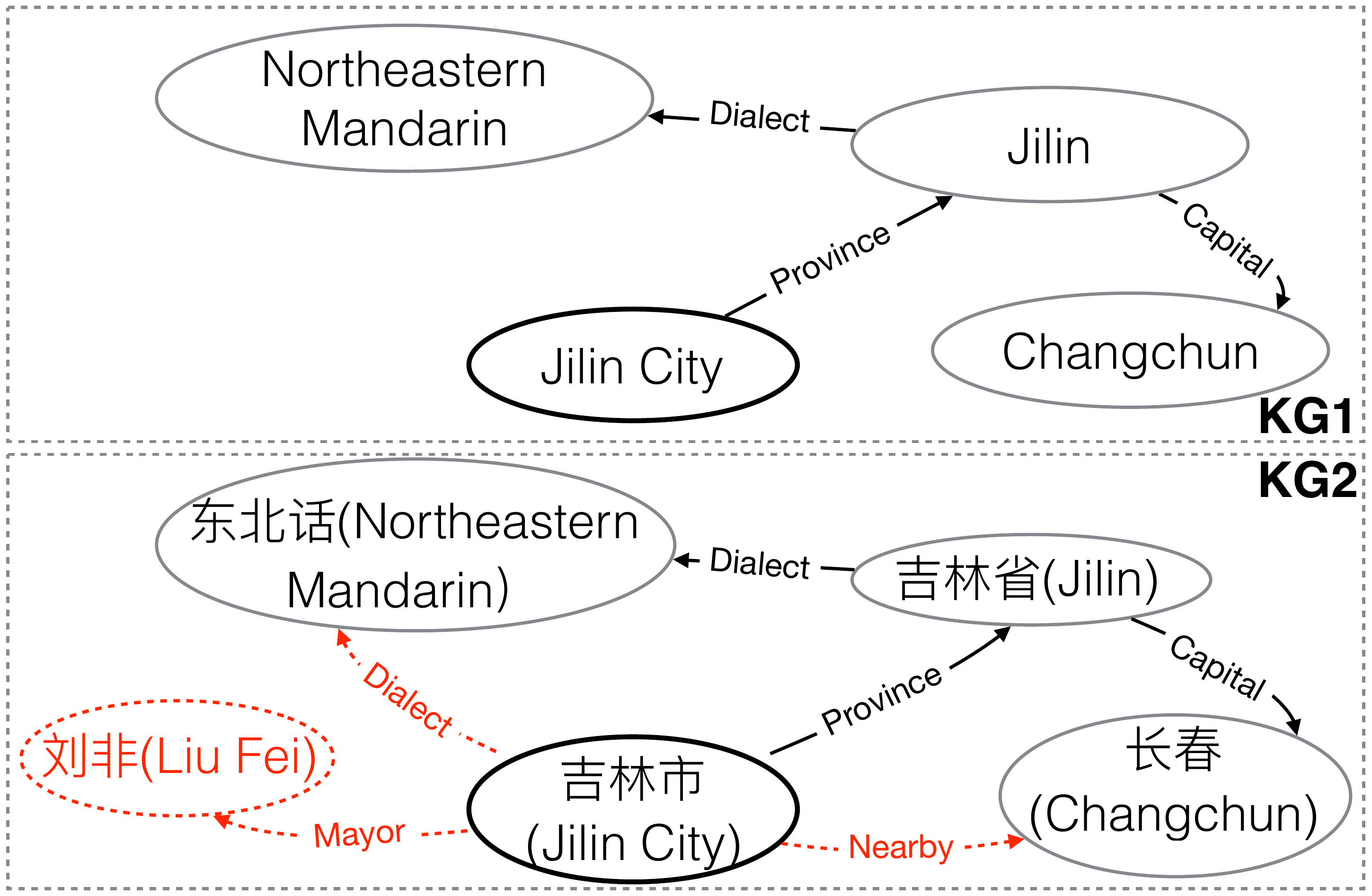}}
	\caption{Illustration of the structural differences (dashed lines and ellipse) between different KGs.}
	\label{fig:example}
\end{figure}

It is non-trivial to align different KGs due to their distinct surface forms, which makes the symbolic based methods~\cite{suchanek2011paris} not always effective. Instead, recent work utilizes general KG embedding methods (e.g., TransE~\cite{bordes2013translating}) and align equivalent entities into a unified vector space based on a few seed alignments~\cite{chen2016multilingual,sun2017cross,zhu2017iterative,chen2018co,sun2018bootstrapping,wang2018cross}. The assumption is that entities and their counterparts in different KGs should have similar structures and thus similar embeddings. However, alignment performance is unsatisfactory mainly due to the following challenges:

\vpara{Heterogeneity of Structures} Different KGs usually differ a lot, and may mislead the representation learning and the alignment information from seeds. Take the entity \textit{Jilin City} as an example (Figure~\ref{fig:example}), KG1 and KG2 present its subgraphs derived from English and Chinese Wikipedia, respectively. Since it is a Chinese city, KG2 is more informative than KG1 (denoted by dashed lines and ellipse), such as the relations of \textit{Dialect} and \textit{Nearby}, and the entity \textit{Liu Fei} through relation \textit{Mayor}. Clearly, the province \textit{Jilin} in KG1 and \textit{Jilin City} in KG2, which are incorrect alignment, are more probable close in the vector space, because they have more similar structures (e.g., \textit{Northeastern Mandarin} and \textit{Changchun}). What's worse, this incorrect alignment shall spread further over the graph.

\vpara{Limited Seed Alignments} Recent efforts based on general embedding methods heavily rely on existing alignments as training data, while seed alignments are usually insufficient~\cite{chen2016multilingual} for high-quality entity embeddings. \citet{wang2018cross} introduces Graph Convolution Network (GCN)~\cite{kipf2016semi} to enhance the entity embeddings by modeling structural features, but fails to consider structural heterogeneity.

To address the issues, we propose to perform KG inference and alignment jointly to explicitly reconcile the structural difference between different KGs, and utilize a graph-based model to make better use of seed alignment information. The basic idea of structural reconciliation is to complete missing relations and prune exclusive entities. As shown in Figure~\ref{fig:example}, to reconcile the differences of \textit{Jilin City}, it is necessary to complete the missing relations \textit{Dialect} and \textit{Nearby} in KG1, and filter out entity \textit{Liu Fei} exclusive in KG2. The asymmetric entities and relations are caused not only by the incompleteness nature of KG, but also from their different demands.


In this paper, we propose a novel \textbf{Mu}lti-channel \textbf{G}raph \textbf{N}eural \textbf{N}etwork model MuGNN, which can encode different KGs to learn alignment-oriented embeddings. For each KG, MuGNN utilizes different channels towards KG completion and pruning, so as to reconcile two types of structural differences: missing relations and exclusive entities. Different channels are combined via pooling techniques, thus entity embeddings are enhanced with reconciled structures from different perspectives, making utilization of seed alignments effectively and efficiently. Between KGs, each channel transfers structure knowledge via shared parameters.

Specifically, for KG completion, we first employ AMIE+~\cite{galarraga2015fast} on each KG to induce rules, then transfer them between KGs towards consistent completion. Following Graph Attention Network (GAT)~\cite{velivckovic2018graph}, we utilize KG self-attention to weighting relations for GNN channels. For KG pruning, we design cross-KG attention to filter out exclusive entities by assigning low weights to corresponding relations. We summarize the main contributions as follows:

\begin{itemize}
	\item We propose a novel Multi-channel GNN model MuGNN that learns alignment-oriented embeddings by encoding graphs from different perspectives: completion and pruning, so as to be robust to structural differences.

	\item We propose to perform KG inference and alignment jointly, so that the heterogeneity of KGs are explicitly reconciled through completion by rule inference and transfer, and pruning via cross-KG attention.

	\item We perform extensive experiments on five publicly available datasets for entity alignment tasks, and achieve significant improvements of 5\% Hits@1 on average. Further ablation study demonstrates the effectiveness of our key components.
\end{itemize}

\begin{figure*}[htb]
	\centerline{\includegraphics[width=0.85\textwidth]{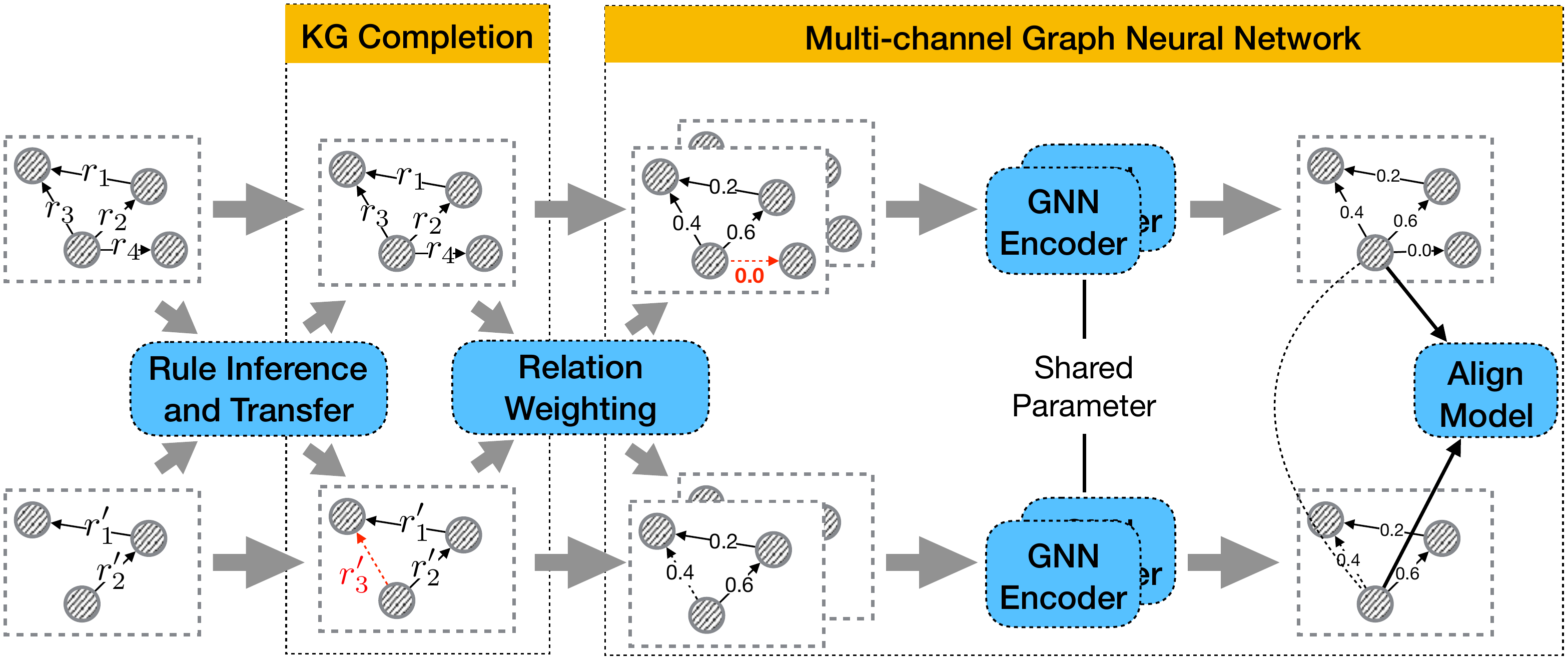}}
	\caption{Framework. Rectangles denote two main steps, and rounded rectangles denote the key components of the corresponding step. After rule inference and transfer, we utilize rules to complete each KG, denoted by dashed lines $r'_3$. Through relation weighting, we obtain multiple weighted graphs for different GNN channels, in which relation $r_4$ is weighted to $0.0$ that prunes exclusive entities. These channels are combined as the input for align model for alignment-oriented KG embeddings.}
	\label{fig:framework}
\end{figure*}

\section{Preliminaries and Framework}
\subsection{Preliminaries}

\para{KG} is a directed graph $G = (E, R, T)$ involving a set of entities $E$, relation types $R$, and triplets $T$. Each triplet $t=(e_i,r_{ij},e_j)\in T$ denotes that head entity $e_i$ is related to tail entity $e_j$ through relation $r_{ij}\in R$.

\para{Rule knowledge} $\mathcal{K}=\{k\}$ can be induced from KG, e.g., in the form of $\forall x,y\in E:(x,r_s,y)\Rightarrow (x,r_c,y)$, stating that two entities might be related through $r_c$ if they are related through $r_s$. The left side of the arrow is defined as premise, and the right side is a conclusion. We denote rule as $k=(r_c|r_{s1},\cdots,r_{sp})$ consisting of one or multiple $|p|$ premises and only one conclusion.

\para{Rule Grounding} is to find suitable triplets satisfying the premise-conclusion relationship defined by rules. For rule $k$, we denote one of its grounds as $g(k)=(t_c|t_{s1},\cdots,t_{sp})$ including $|p|+1$ triplets. The triplets satisfies: $t_{s1}\wedge\cdots \wedge t_{sp}\Rightarrow t_c$, where $\wedge$ is the logical conjunction that plays a similar role as `and'. Other compositions include disjunction $\vee$ (similar to `or') and negation $\neg$ (similar to `not'). For example, given a rule \textit{bornIn}$(x,y)$ $\wedge$ \textit{cityOf}$(y,z)$ $\Rightarrow$ \textit{nationality}$(x,z)$, we ground it in a KG, and obtain : \textit{bornIn}(\textit{Obama}, \textit{Hawaii}) $\wedge$ \textit{cityOf}(\textit{Hawaii}, \textit{United States}) $\Rightarrow$ \textit{nationality}(\textit{Obama}, \textit{United States}). We use $\mathcal{G}(k)=\{g(k)\}$ to denote all groundings of rule $k$.

\para{Entity alignment} takes two heterogeneous KGs $G$ and $G' = (E', R', T')$ as input, the goal is to find as many alignments as possible $\mathcal{A}_e = \{(e, e') \in E \times E'|e \leftrightarrow e'\}$ for which an equivalent relation $\leftrightarrow$ holds between $e$ and $e'$. That is, $e$ and $e'$ are in different KGs, but denote the same thing. As shown in Figure~\ref{fig:example}, \textit{Jilin City} in English Wikipedia (i.e., KG1) and in Chinese Wikipedia (i.e., KG2) has different structures, but denote the same Chinese city. Normally, some prior alignments of entities $\mathcal{A}_e^s$ and relations $\mathcal{A}_r^s = \{(r, r') \in R \times R'|r \leftrightarrow r'\}$ can be easily obtained manually or by simple lexicon-based methods (e.g., entity title translation), namely seed alignments (seed for short). We use bold-face letters to denote the vector representations of the corresponding terms throughout of the paper.

\subsection{Framework}
MuGNN aims at learning alignment-oriented KG embeddings for entity alignment. It introduces KG inference and transfer to explicitly complete KGs, and utilizes different relation weighting schemes: KG self-attention and cross-KG attention, to encode KGs robustly. As shown in Figure~\ref{fig:framework}, there are two main steps in our framework:

\vpara{KG Completion} aims at reconciling the structural differences by completing the missing relations. It not only induces rules by using a popular rule mining system AMIE+~\cite{galarraga2015fast}, but also transfers them into each other based on seed aligned relations between KGs. Rule transferring is based on the assumption that knowledge can be generalized into various KGs, no matter in which languages or domains.

\vpara{Multi-channel Graph Neural Network} is to encode each KG through different channels. The channels enhance the entity embeddings from different perspectives: towards completion and pruning, so that the entities and their counterparts have similar structures. MuGNN contains three main components: (1) relation weighting that generates weight matrix for each KG according to two schemes: KG self-attention and cross-KG attention. Each type of attention refers to a GNN channel that shares parameters between KGs for structural knowledge transfer; (2) GNN encoder to model the entire graph features by improving entity embeddings with its neighbors, thus the seed alignment information shall be propagated over the entire graph; We combine the outputs of GNN encoders in different channels via pooling techniques as the input of (3) Align model, which embeds two KGs into a unified vector space by pushing the aligned entities (and relations) of seeds together.

\section{KG Completion}
In this section, we introduce how to utilize rule knowledge to explicitly complete KG, which first infers rules from each KG, then transfers these rules between KGs based on knowledge invariant assumption, and finally grounds rules in each KG for consistent completion.

\subsection{Rule Inference and Transfer}
Since the acquirement of rule knowledge is not our focus in this paper, we utilize AMIE+~\cite{galarraga2015fast}, a modern rule mining system, to efficiently find Horn rules from large-scale KG, such as \textit{marriedTo}$(x,y)$ $\wedge$ \textit{liveIn}$(x,z)$ $\Rightarrow$ \textit{liveIn}$(y,z)$. Its source code is available online\footnote{\url{https://www.mpi-inf.mpg.de/}}.

Formally, given two KGs $G$ and $G'$, we first mine rules separately and obtain two sets of rule knowledge $\mathcal{K}$ and $\mathcal{K}'$. These rule knowledge are quite different since KGs are constructed to meet different demands of applications or languages. Although they can be used to complete their own KGs separately, we further transfer the two sets of rules into each other through \textbf{Knowledge Invariant Assumption}:

\textit{Knowledge has universality no matter in which languages or domains.}

Given aligned relations $\mathcal{A}_r^s$ and a rule $k\in\mathcal{K}$, we replace all relations involved in the rule $k=(r_c|r_{s1},\cdots,r_{sp})$ with its counterparts if there are $(r_c,r_c'),(r_{si},r_{si}')\in \mathcal{A}_r^s,i=1,\cdots,p$. Thus, we obtain such a rule $k'=(r_c'|r_{s1}',\cdots,r_{sp}')$ and add it to $\tilde{\mathcal{K}}'=\mathcal{K}'\cup k'$ if $k'\notin \mathcal{K}'$. Real examples of transferred rules can be found in experiments. Note that there may be no transfered rules if aligned relations can not be found $\mathcal{A}_r^s=\emptyset$.

\subsection{Rule Grounding}
We now ground each rule sets on the corresponding KG for completion, which not only accelerates the efficiency of align model through denser KG for propagation, but also adds extra constraints that is helpful for high-quality entity embedding learning.

Take KG $G$ as an example, given a rule $k\in\mathcal{K}$, we collect its grounds that the premise triples can be found in the KG, but not the conclusion triplet: $\mathcal{G}(k)=\{g(k)|t_{s1},\cdots,t_{sp}\in T,t_c\notin T\}$. Thus, we add all conclusion triples into the KG $\tilde{G}=G\cup t_c, t_c\in\mathcal{G}(k)$. Similarly, we can complete KG $G'$ to $\tilde{G}'$.

As shown in Figure~\ref{fig:example}, we obtain the rule \textit{province}$(x,y)$ $\wedge$ \textit{dialect}$(y,z)$ $\Rightarrow$ \textit{dialect}$(x,z)$ from the informative KG2, then transfer it to KG1 based on the aligned relation \textit{province} and \textit{dialect}. Thus, in KG1, we find the suitable triplets \textit{province}(\textit{Jilin City}, \textit{Jilin}) $\wedge$ \textit{dialect}(\textit{Jilin}, \textit{Northeastern Mandarin}), thus obtain a new triplet \textit{dialect}(\textit{Jilin City}, \textit{Northeastern Mandarin}).

It is worth noting that the inferred rules do not hold in all cases, and maybe we can consider the confidence value for each grounding. We leave it in future work.

\section{Multi-Channel Graph Neural Network}
In this section, we describe the three main components involved in MuGNN to encode different graphs towards alignment-oriented embedding learning: relation weighting, multi-channel GNN encoder and align model.

\subsection{Relation Weighting}
Relation weighting is to generate weighted connectivity matrix $A$ based on a graph $G$ as the input structural features of GNN encoder, which will be detailed later. Each element $a_{ij}$ in the matrix denote the weighted relation between $e_i$ and $e_j$.

As mentioned in Section~\ref{sec:intro}, there are two types of structure differences: the missing relations due to the incompleteness nature of KG, and the exclusive entities caused by different construction demands of applications or languages. We utilize two channels of GNN encoder for each KG, so as to reconcile the two types of differences separately. That is, we generate two adjacency matrices for each channel: $A_1$ based on KG self-attention and $A_2$ based on cross-KG attention. Next, we will describe how to compute each element $a_{ij}$ in $A_1$ and $A_2$. Similarly, we can obtain $A'_1$ and $A'_2$ for KG $G'$.

\subsubsection*{KG Self-Attention}
KG self-attention aims at making better use of seed alignments based on the KG structure itself. This component selects informative neighbors according to the current entity and assigns them with high weights. Following GAT~\cite{velivckovic2018graph}, we define the normalized element $a_{ij}$ in $A_1$ representing the connectivity from entity $e_i$ to $e_j$ as follows:

\begin{equation}
	a_{ij} = softmax(c_{ij}) = \frac{exp(c_{ij})}{\sum_{e_k \in N_{e_i}\cup{e_i}} exp(c_{ik})}
\end{equation}
where $e_k \in N_{e_i} \cup \{e_i\}$ denotes neighbors of $e_i$ with self-loop, and $c_{ij}$ is the attention coefficient measuring the importance of $e_i$ to $e_j$ and is calculated by an attention function $attn$ as follows:

\begin{equation}
	\begin{split}
		c_{ij} &= attn(\mathbf{W} \mathbf{e_i}, \mathbf{W} \mathbf{e_j}) \\
		&= LeakyReLU(\mathbf{p}[\mathbf{W} \mathbf{e_i} || \mathbf{W} \mathbf{e_j}])
	\end{split}
\end{equation}
where $||$ indicates vector concatenation, $\mathbf{W}$ and $\mathbf{p}$ are trainable parameters.

\subsubsection*{Cross-KG Attention}
Cross-KG Attention aims at modeling the common subgraph of two KGs as structural features towards consistency. It prunes exclusive entities by assigning lower weights for corresponding relations that have no counterparts in another KG. We define the $a_{ij}$ in $A_2$ as follows:

\begin{equation}
	a_{ij} = \underset{r\in R,r'\in R'}{\max} \mathbf{1}((e_i,r,e_j)\in T) sim(r, r')
\end{equation}
where $\mathbf{1}(\cdot)$ indicates $1$ if holds true, otherwise 0. $sim(\cdot)$ is a similarity measure between relation types and is defined as inner-product $sim(r, r')=\mathbf{r}^T\mathbf{r}'$. Thus, $a_{ij}$ is to find the best mapping between two KGs, and shall be $0$ if there is no such relation types for exclusive entities.

\subsection{Multi-Channel GNN Encoder}

GNN is a type of neural network model that deals with graph-structured data, the main idea of which is similar to a propagation model: to enhance the features of a node (i.e., entity) according to its neighbor nodes. Thus, we may stack multiple $L$ layers of GNNs to achieve further propagation.

One of its variants is based on spectral graph convolutions, such as GCN~\cite{kipf2016semi}. Every GNN encoder takes the hidden states of node representations in the current layer as inputs, and computes new node representations as:

\begin{equation} \label{eq:gat}
	\text{GNN}(A,H,W) = \sigma(\mathbf{A} \mathbf{H} \mathbf{W})
\end{equation}
where $\mathbf{A}$ is an adjacency matrix showing the connectivity between nodes, $\mathbf{H}$ is the current node representations, $\mathbf{W}$ is the learned parameters, and $\sigma$ is the activation function chosen as $ReLU(\cdot) = max(0, \cdot)$.

Inspired by the multi-head attention networks~\cite{velivckovic2018graph}, we use the two above-mentioned strategies to calculate connectivity matrices as different channels to propagate information from different aspects and aggregate them with a $Pooling$ function. As for our multi-channel GNN encoder, it is built by stacking multiple GNN encoder defined as:

\begin{equation} \label{eq:gat}
	\begin{split}
		\text{MultiGNN}(H^l;&A_1,\cdots,A_c)= \\
		&\text{Pooling}(H^{l+1}_1,\cdots,H^{l+1}_c)
	\end{split}
\end{equation}
where $c$ is the number of the channels, $A_i$ is the connectivity matrices in the $i_{th}$ channel, and $H^{l+1}_i$ is the computed hidden states in the $(l+1)_{th}$ layer and $i_{th}$ channel, which can be formulated as:

\begin{equation} \label{eq:gat}
	\mathbf{H}^{l+1}_i = \text{GNN}(A_i,H^l,W_i)
\end{equation}
where $W_i$ is the weight parameters in the $i_{th}$ channel. Here, we set $i=1,2$ referring to the above two attention schemes. We set $H^0$ as the entity embeddings initialized randomly. In experiments, we select average pooling techniques for $\text{Pooling}$ function due to its superior performance.

We use such multi-channel GNN encoders to encode each KG, and obtain $\mathbf{H}^L$, $\mathbf{H}'^L$ representing the enhanced entity embeddings, where each channel shares parameters $W_1=W'_1$ and $W_2=W'_2$ for structural knowledge transferring.

\subsection{Align Model}
Align model is to embed two KGs into a unified vector space by pushing the seed alignments of entities (and relations) together. We judge whether two entities or two relations are equivalent by the distance between them. The objective of the align model is given as below:

\begin{equation}
	{\footnotesize
		\begin{split}
			\mathcal{L}_a &= \sum_{(e, e^{'}) \in \mathcal{A}_e^s} \sum_{(e_{-}, e_{-}^{'}) \in {\mathcal{A}_e^s}^{-}}[d(e, e^{'}) + \gamma_1 - d(e_{-}, e_{-}^{'})]_{+} + \\
			& \sum_{(r, r^{'}) \in \mathcal{A}_r^s} \sum_{(r_{-}, r_{-}^{'}) \in {\mathcal{A}_r^s}^{-}}[d(r, r^{'}) + \gamma_2 - d(r_{-}, r_{-}^{'})]_{+}
		\end{split}}
\end{equation}
where $[\cdot]_{+} = max\{0, \cdot\}$ represents the maximum between 0 and the input, $d(\cdot) = ||\cdot||_2$ is the distance measure chosen as L2 distance, ${\mathcal{A}_e^s}^{-}$ and ${\mathcal{A}_r^s}^{-}$ represents for the negative pair set of $\mathcal{A}_e^s$ and $\mathcal{A}_r^s$, respectively, and $\gamma_1 > 0$ and $\gamma_2 > 0$ are margin hyper-parameters separating positive and negative entity and relation alignments. During the experiments, by calculating cosine similarity, we select 25 entities closest to the corresponding entity in the same KG as negative samples~\cite{sun2018bootstrapping}. Negative samples will be re-calculated every 5 epochs.

\subsubsection*{Rule Knowledge Constraints}
Since we have changed the KG structure by adding new triplets (i.e., grounded rules), we also introduce the triplet loss to hold the grounded rules as valid in the unified vector space.

Taking KG $G$ as an example, following~\newcite{guo2016jointly}, we define the loss function as follows:

\begin{equation}
	\begin{split}
		\mathcal{L}_r = & \underset{g^+\in\mathcal{G}(\mathcal{K})}{\sum} \underset{g^-\in\mathcal{G}^-(\mathcal{K})}{\sum} [\gamma_r-I(g^+)+I(g^-)]_+ \\
		+ & \underset{t^+\in T}{\sum} \underset{t^-\in T^-}{\sum} [\gamma_r-I(t^+)+I(t^-)]_+
	\end{split}
\end{equation}
where $g$ is short for rule grounding $g(k)$, $\mathcal{G}(\mathcal{K})$ and $T$ denote all rule grounds and all triplets. $\mathcal{G}^-(\mathcal{K})$ and $T^-$ are negative sample sets obtained by replacing one of the involved entity using nearest sampling~\cite{sun2018bootstrapping}. $I(\cdot)$ is the true value function for triplet $t$:

\begin{equation}
	I(t)=1-\frac{1}{3\sqrt{d}}||\mathbf{e}_i+\mathbf{r}_{ij}-\mathbf{e}_j||_2
\end{equation}
or for grounding $g=(t_c|t_{s1},\cdots,t_{sp})$, which is recursively calculated by:

\begin{equation}
	\begin{split}
		I(t_s) = I(t_{s1}\wedge t_{s2})=I(t_{s1})\cdot I(t_{s2}) \\
		I(t_s \Rightarrow t_c)=I(t_s)\cdot I(t_c) - I(t_s) + 1
	\end{split}
\end{equation}
where $d$ is the embedding size. Similarly, we obtain the loss $\mathcal{L}'_r$ for KG $G'$. Thus, the overall loss function for multi-channel GNN is as follows:

\begin{equation}
	\mathcal{L} = \mathcal{L}_a + \mathcal{L}'_r + \mathcal{L}_r
\end{equation}

\section{Experiment}
In this section, we conduct experiments on five publicly available datasets involving both different language pairs and sources. We further investigate the key components of MuGNN and analyze how the knowledge inference and transfer mechanism contribute to KG alignment.

\subsection{Experiment Settings}

\para{Datasets} Following~\citet{sun2017cross, sun2018bootstrapping}, we conduct experiments on benchmark datasets DBP15K and DWY100K. DBP15K contains three cross-lingual datasets: DBP{\tiny ZH-EN}(Chinese to English), DBP{\tiny JA-EN} (Japanese to English), and DBP{\tiny FR-EN} (French to English). All the above datasets are extracted from multilingual DBpedia and include 15,000 entity pairs as seed alignments. DWY100K consists of two large-scale cross-resource datasets: DWY-WD (DBpedia to Wikidata) and DWY-YG (DBpedia to YAGO3). Each dataset includes 100,000 alignments of entities in advance. As for the seed alignments of relations, we employ the official relation alignment list published by DBpedia for DWY100K. As for DWY-YG, we manually align the relations because there are only a small set of relation types (31) in YAGO3. The statistics\footnote{$|\mathcal{A}_r^s|$ denotes the number of seed alignments of relations.} is listed in Table \ref{tab:dataset}.

\begin{table}[htbp]
	\centering
	\setlength{\tabcolsep}{1.0pt}
	\small
	\renewcommand{\arraystretch}{1.0}
	\begin{tabular}{ccccc}
		\toprule
		\textbf{Datasets}      & $\mathbf{|\mathcal{A}_r^s|}$      & \#\textbf{Relation} & \#\textbf{Entity}  & \#\textbf{Triple}  \\
		\midrule
		\textbf{DBP{\tiny ZH}} & \multirow{2}[2]{*}{891 } & 2,830    & 66,469  & 153,929 \\
		\textbf{DBP{\tiny EN}} &                          & 2,317    & 98,125  & 237,674 \\
		\midrule
		\textbf{DBP{\tiny JA}} & \multirow{2}[2]{*}{582 } & 2,043    & 65,744  & 164,373 \\
		\textbf{DBP{\tiny EN}} &                          & 2,096    & 95,680  & 233,319 \\
		\midrule
		\textbf{DBP{\tiny FR}} & \multirow{2}[2]{*}{75 }  & 1,379    & 66,858  & 192,191 \\
		\textbf{DBP{\tiny EN}} &                          & 2,209    & 105,889 & 278,590 \\
		\midrule
		\textbf{DWY{\tiny DB}} & \multirow{2}[2]{*}{62 }  & 330      & 100,000 & 463,294 \\
		\textbf{DWY{\tiny WD}} &                          & 220      & 100,000 & 448,774 \\
		\midrule
		\textbf{DWY{\tiny DB}} & \multirow{2}[2]{*}{24 }  & 302      & 100,000 & 428,952 \\
		\textbf{DWY{\tiny YG}} &                          & 31       & 100,000 & 502,563 \\
		\bottomrule
	\end{tabular}%
	\caption{Statistics of DBP15K and DWY100k.}
	\label{tab:dataset}%
\end{table}%

For each dataset, we employ AMIE+ for rule mining by setting the max number of premise as $p=2$ and PCA confidence not less than $0.8$. The statistical results of rules, transferred rules (Tr.Rule for short), ground triples and ground triples based on transferred rules (Tr.ground for short) are exhibited in Table \ref{tab:ruleset}.

\begin{table}[htbp]
	\centering
	\small
	\setlength{\tabcolsep}{1.0pt}
	\renewcommand{\arraystretch}{1.0}
	\begin{tabular}{ccccc}
		\toprule
		\textbf{Datasets}      & \#\textbf{Rule} & \#\textbf{Tr.Rule} & \#\textbf{Ground} & \#\textbf{Tr.ground} \\
		\midrule
		\textbf{DBP{\tiny ZH}} & 2,279       & 1,058       & 46,959      & 19,278      \\
		\textbf{DBP{\tiny EN}} & 1,906       & 578         & 78,450      & 24,018      \\
		\midrule
		\textbf{DBP{\tiny JA}} & 1,440         & 651         & 61,733      & 25,337      \\
		\textbf{DBP{\tiny EN}} & 1,316       & 259         & 77,614      & 17,838      \\
		\midrule
		\textbf{DBP{\tiny FR}} & 1,263       & 25          & 77,342      & 1,527       \\
		\textbf{DBP{\tiny EN}} & 1,252       & 12          & 75,338      & 1,364       \\
		\midrule
		\textbf{DWY{\tiny DB}} & 843         & 40          & 281,271     & 13,136      \\
		\textbf{DWY{\tiny WD}} & 630         & 51          & 184,010     & 56,373      \\
		\midrule
		\textbf{DWY{\tiny DB}} & 503         & 4           & 277,031     & 92,923      \\
		\textbf{DWY{\tiny YG}} & 39          & 16          & 129,334     & 10,446      \\
		\bottomrule
	\end{tabular}
	\caption{Statistics of KG inference and transfer.}
	\label{tab:ruleset}
\end{table}%

\begin{table*}[htbp]
	\centering
	\small
	\setlength{\tabcolsep}{2.0pt}
	\begin{tabular}{cccccccccccccccc}
		\toprule
		\multirow{2}[4]{*}{\textbf{Methods}} & \multicolumn{3}{c}{\textbf{DBP\tiny{ZH-EN}}} & \multicolumn{3}{c}{\textbf{DBP\tiny{JA-EN}}} & \multicolumn{3}{c}{\textbf{DBP\tiny{FR-EN}}} & \multicolumn{3}{c}{\textbf{DBP-WD}} & \multicolumn{3}{c}{\textbf{DBP-YG}} \\
		\cmidrule{2-16}             & \textbf{H@1}            & \textbf{H@10}          & \textbf{MRR}            & \textbf{H@1}            & \textbf{H@10}           & \textbf{MRR}            & \textbf{H@1}            & \textbf{H@10}           & \textbf{MRR}      & \textbf{H@1}            & \textbf{H@10}           & \textbf{MRR}            & \textbf{H@1}            & \textbf{H@10}           & \textbf{MRR}            \\
		\midrule
		\textbf{MTransE}                     & .308          & .614          & .364          & .279          & .575          & .349          & .244          & .556          & .335          & .281          & .520          & .363          & .252          & .493          & .334          \\
		\textbf{JAPE}                        & .412          & .745          & .490          & .363          & .685          & .476          & .324          & .667          & .430          & .318          & .589          & .411          & .236          & .484          & .320          \\
		\textbf{AlignEA}                     & .472          & .792          & .581          & .448          & .789          & .563          & .481          & .824          & .599          & .566          & .827          & .655          & .633          & .848          & .707          \\
		\textbf{GCN-Align}                   & .413          & .744          & .549          & .399          & .745          & .546          & .373          & .745          & .532          & .506          & .772          & .600          & .597          & .838          & .682          \\
		\midrule
		\textbf{MuGNN w/o} $\mathbf{\mathcal{A}_r^s}$ & .479          & .833          & .597          & .487          & .851          & .604          & \textbf{.496} & .869          & \textbf{.621} & .590          & .887          & .693          & .730          & .934          & .801          \\
		\textbf{MuGNN}                       & \textbf{.494} & \textbf{.844} & \textbf{.611} & \textbf{.501} & \textbf{.857} & \textbf{.621} & .495          & \textbf{.870} & \textbf{.621} & \textbf{.616} & \textbf{.897} & \textbf{.714} & \textbf{.741} & \textbf{.937} & \textbf{.810} \\
		\bottomrule
	\end{tabular}%
	\caption{Overall performance.}
	\label{tab:performance}%
\end{table*}%
  
\vpara{Baselines}
To investigate MuGNN's ability on entity alignment, we select four competitive baselines including three translation based models and one graph-based model for comparison. \textbf{MTransE}~\cite{chen2016multilingual} trains independent embedding of knowledge graph with TransE, and assigns the entity pairs in seed alignments with similar embeddings by minimizing their Euclidean distances. \textbf{JAPE}~\cite{sun2017cross} learns the representation of entities and relations from different KGs in a unified embedding space. It takes advantage of attribute triples to capture homogeneous entity properties cross KGs. \textbf{GCN-Align}~\cite{wang2018cross} employs Graph Convolution Networks to construct entity representation by propagating information from the neighborhood.  \textbf{AlignEA}~\cite{sun2018bootstrapping} swaps aligned entities in triples to calibrate the embedding of KGs in a unified embedding space. AlignEA is the up to date non-iterative state of the art model.

\vpara{Training Details}
Following~\citet{sun2017cross, sun2018bootstrapping}, we split 30\% of entity seed alignments as training data and left the remaining data for testing. By convention, Hits@N and Mean Reciprocal Rank are used as evaluation metrics. Hits@N indicates the percentage of the targets that have been correctly ranked in top N (H in Table~\ref{tab:performance} for short). MRR is the average of the reciprocal of the rank results. Higher Hits@N and MRR refer to higher performance.

To make a fair comparison, we set embedding size to 128 for MuGNN and all baselines. All graph models stack two layers of GNN. We utilize Adagrad~\cite{duchi2011adaptive} as the optimizer. For the margins in MuGNN, we empirically set $\gamma_1=1.0$ and $\gamma_2=1.0$. We set $\gamma_r=0.12$ to ensure rule knowledge constraints have less impact than the alignment model. Other hyperparameters are chosen by running an exhaustively search over the following possible values: learning rate in $\{0.1, 0.01, 0.001\}$, L2 in $\{0.01, 0.001, 0.0001\}$, dropout in $\{0.1, 0.2, 0.5\}$. The optimal configuration of MuGNN for entity alignment is: learning rate$=0.001$, L2$=0.01$, dropout $= 0.2$. We implement MuGNN with PyTorch-1.0. The experiments are conducted on a server with two 6-core Intel Xeon E5-2620 v3@2.40ghz CPUs, two GeForce GTX TITAN X and 128 GB of memory. 500 epochs cost nearly one hour.

\subsection{Overall Performance}

Table \ref{tab:performance} shows the experimental results on DBP15K and DWY100K. In general, MuGNN significantly outperforms all baselines regarding all metrics, mainly because it reconciles the structural differences by two different schemes for KG completion and pruning, which are thus well modeled in multi-channel GNN.

More specifically, on three small-scale cross-lingual datasets, the average gains of MuGNN regarding Hits@1, Hits@10 and MRR are 3\%, 6\%, and 4\%, respectively. While on large-scale datasets, MuGNN achieves significant improvements (8\%, 8\% and 8\% regarding Hits@1, Hits@10 and MRR, respectively). This is mainly because the large-scale datasets (e.g., DBP-YG) provide more prior knowledge (more than 3.5 facts per entity v.s. less than 2.5 facts in DBP15K) for rule mining, thus our proposed method has more capability in reconciling the structural differences between KGs, and makes better use of seed alignments.

\begin{figure*}[htb]
	\centerline{\includegraphics[width=0.9\textwidth]{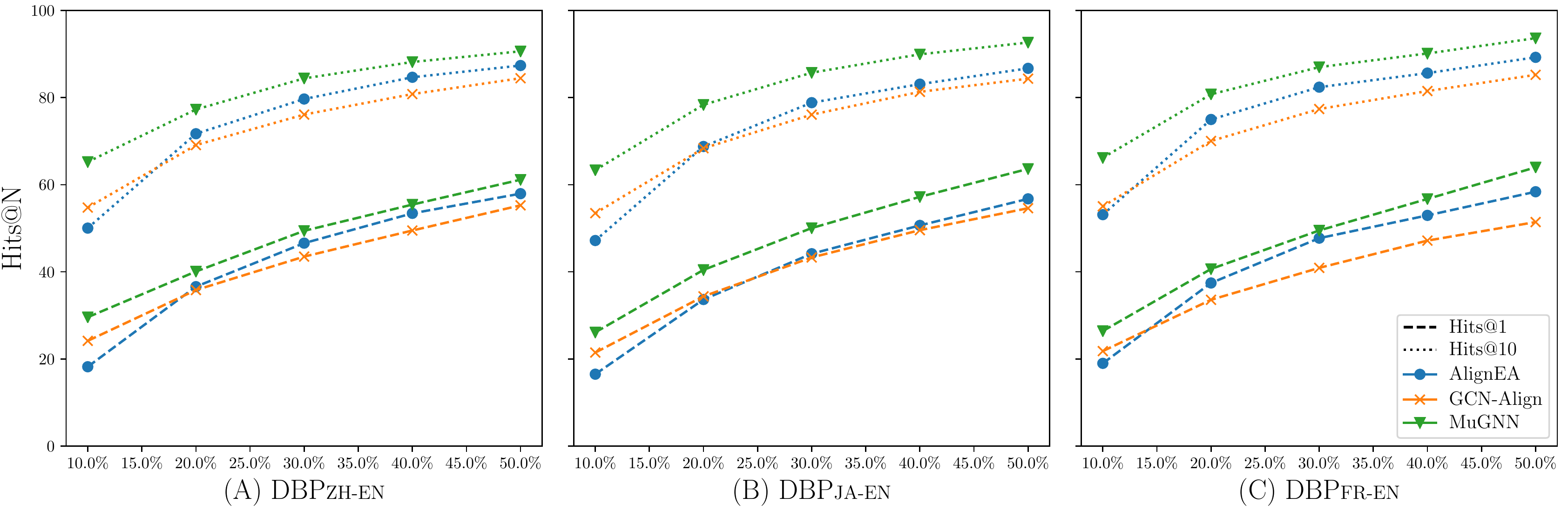}}
	\caption{Sensitivity to entity seed alignments (x-axis: proportion of seed alignments used for training).}
	\label{fig:seed}
\end{figure*}

Since some methods do not rely on seed alignments of relations, we also test MuGNN without them, marked as MuGNN w/o $\mathcal{A}_r^s$. This also implies that we have no transferred rules between KGs. We can see that our method still performs competitively, and even achieves the best Hits@1 and MRR on {DBP\tiny{FR-EN}}. This is because the culture difference between French and English is much smaller than that between Chinese/Japanese and English, thus there is only a few exclusive rules mined from each KG, which can be transferred towards consistent completion (25 and 12 pieces of rules transferred between two KGs, as shown in Table~\ref{tab:ruleset}).

We also observe that GNN-based method (i.e., GCN-Align) performs better than translation-based methods except AlignEA. To better understand their advantages and disadvantages, we further conduct ablation study as follows.

\subsection{Impact of Two Channels and Rule Transfer}

\begin{figure}[htb]
	\centerline{\includegraphics[width=0.5\textwidth]{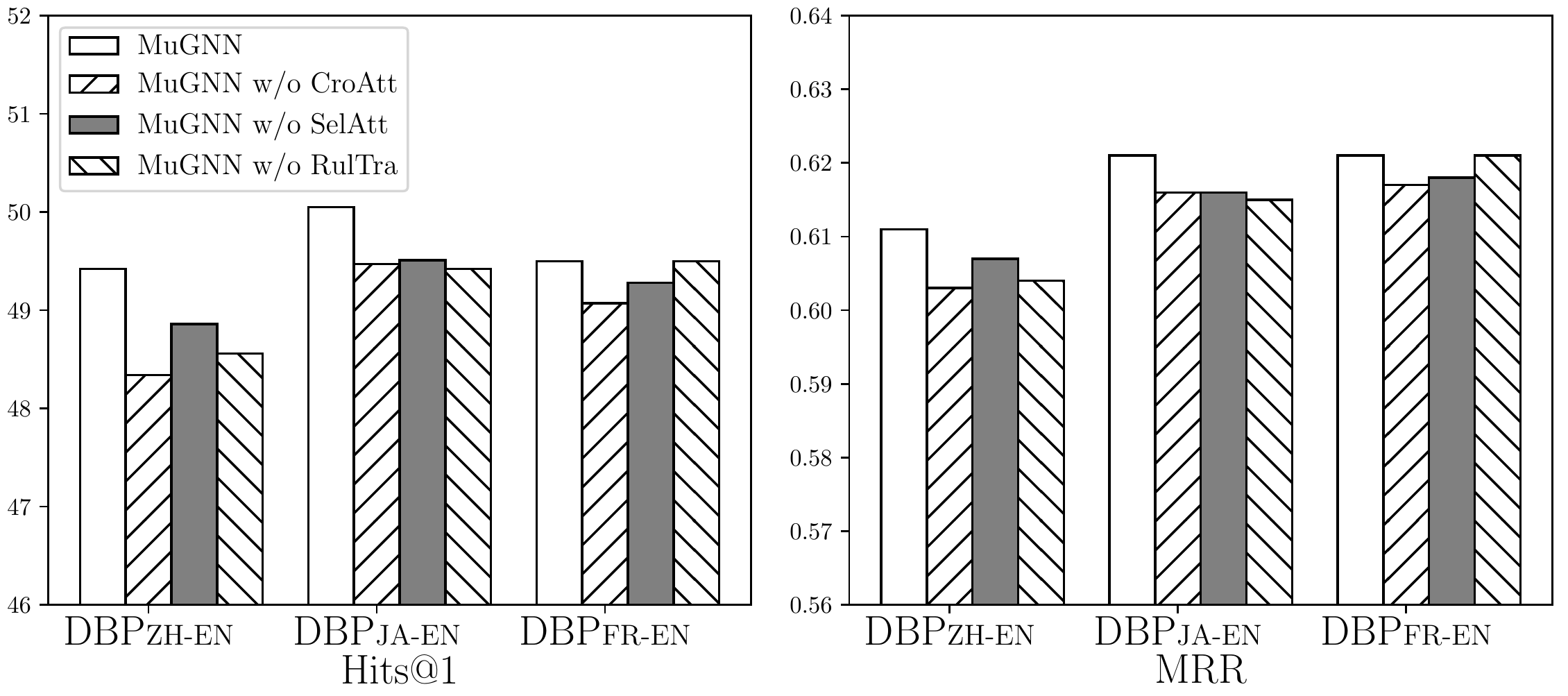}}
	\caption{Impact of two channels and rule transfer.}
	\label{fig:ablation}
\end{figure}

\begin{table*}[htp]
	\small
	\centering
	\begin{tabular}{p{15.5cm}}
		\toprule
		\emph{(U.S., leaderTitle, U.S.President) $\wedge$ (U.S.Secretary of State, reports to, U.S.President) $\Rightarrow$ (U.S.Secretary of State, seat, U.S.)} \\
		\emph{(Chiang Kaishek,party,Kuomingtang) $\wedge$ (Chiang Weikuo,president,Chiang Kaishek) $\Rightarrow$ (Chiang Weikuo,party,Kuomintang)}                \\
		\bottomrule
	\end{tabular}
	\caption{Examples of groundings of transferred rules.}
	\label{tab:gr_rule}
\end{table*}

The core components of MuGNN involve two channels based on KG self-attention and cross-KG attention, and rule transfer towards consistent completion based on knowledge invariant assumption. We thus remove them from our model to to investigate their impacts to reconcile the structural differences, marked as MuGNN w/o SelAtt, MuGNN w/o CroAtt and MuGNN w/o RulTra.

As shown in Figure \ref{fig:ablation}, there is a performance drop in MuGNN w/o SelAtt and MuGNN w/o CroAtt as compared to MuGNN, which demonstrates the effectiveness of both channels. Specifically, the performance decrease more with the loss of cross-KG attention channel than that of KG self-attention, which implies the importance of utilizing cross-KG information for entity alignment. As for rule transfer, we can see that in most cases, it contributes much in performance. However, the performance difference between MuGNN and MuGNN w/o RulTra is negligible on DBP{\tiny FR-EN}. The reason is that the ground rule triple amounts for French and English datasets are limited (Table \ref{tab:ruleset}), which are less than 1\% of the oracle triples. Therefore, rule transfer cannot provide sufficient cross-graph heterogeneous structure information. As a contrast, DBP{\tiny JA-EN} and DBP{\tiny ZH-EN} provide more than 10k ground-rule triples, which gain decent performance improvements from rule transfer.

\subsection{Impact of Seed Alignments}

To investigate the advantages and disadvantages between GNN-based method and translation-based methods, we test MuGNN, GCN-Align and AlignEA using different size of seed alignments. We gradually increase the proportion of entity seeds from 10\% to 50\%, and we can see the model's sensitivity to seed alignments.

As shown in Figure~\ref{fig:seed}, GNN-based methods perform better than translation-based methods when there is only limited seeds available (10\%), but perform worse along with the increase of seed alignments. This is because graph models can make better of seeds by propagating them over the entire structure, while they suffer from the heterogeneity between KGs due to the GNN's sensitivity to structural differences, which lead to propagation errors aggregation. However, the performance of translation-based methods increases gradually along with the growing seeds since it can implicitly complete KG via knowledge representation learning, such as transE. MuGNN utilizes AMIE+ to explicitly complete two KGs via rule mining and transfer, which reconciles the structural differences; meanwhile, the GNN encoders make better use of seed information via two channels over the graphs.

\subsection{Qualitative Analysis}

We qualitatively analyze how the rule works by presenting the transferred rules and their groundings in Table~\ref{tab:gr_rule}. We can see the rule grounding in the first line indicates a common knowledge in the United States, which thus is easily mined in English KG DBP{\tiny EN}. Meanwhile, we find that such knowledge is missing in DBP{\tiny ZH}, the Chinese KG. By transferring the corresponding rules from DBP{\tiny EN} to DBP{\tiny ZH}, the asymmetric information is smoothed. Corresponding entities in Chinese DBP{\tiny ZH} shall have a similar structure with their counterparts in English DBP{\tiny EN}, thus similar embeddings. That is, MuGNN indeed reconciles structural differences by rule transfer, and learns alignment-oriented embeddings. The second line presents a similar case that transfers a Chinese common rule knowledge into English KG. This demonstrates the effectiveness of rule transfer.

\noindent\textbf{Error Analysis}: As shown in Table \ref{tab:ruleset}, the only 4 rules transfer from YAGO3 to DBpedia are grounded to 92,923 new ground rule triples, which is shocking and not informative. Further investigation finds that the rule \emph{(a, team, b) $\Rightarrow$ (a, affiliation, b)} alone contributes 92,743 ground rule triples. Although the rule is logically correct, it is suspicious such a rule that establishes similar relations between entities would benefit entity alignment. We will deal with such noise in future.

\section{Related Work}

Merging different KGs into a unified one has attracted much attention since it shall benefit many Knowledge-driven applications, such as information extraction~\cite{cao2017bridge,cao2018joint}, question answering~\cite{zhang2015target} and recommendation~\cite{cao2019unifying}.
Early approaches for entity alignment leverage various features to overcome the heterogeneity between KGs, such as machine translation and external lexicons~\cite{suchanek2011paris,wang2013boosting}. Following the success of KG representation learning, recent work embeds entities in different KGs into a low-dimensional vector space with the help of seed alignments~\cite{chen2016multilingual}. However, the limited seeds and structural differences take great negative impacts on the quality of KG embeddings, which performs alignment poorly. JAPE~\cite{sun2017cross} and KDCoE~\cite{chen2018co} introduced attributes or descriptions information to improve entity embeddings, while IPTransE~\cite{zhu2017iterative} and BootEA~\cite{sun2018bootstrapping} enlarged the seed set by selecting predicted alignments with high confidence iteratively.

Clearly, the above strategies can be seen as a general enhancement for most alignment approaches~\cite{sun2018bootstrapping}, thus we focus on improving the alignment performance without any external information and in a non-iterative way. Inspired by~\citet{wang2018cross}, which utilize Graph Convolutional Network (GCN)~\cite{kipf2016semi} to encode the entire KGs, we aim at reconciling the heterogeneity between KGs through completion and pruning, and learn alignment-oriented KG embeddings by modeling structural features from different perspectives via Multi-channel GNNs.

\section{Conclusions}

In this paper, we propose a novel Multi-channel Graph Neural Network model, MuGNN, which learns alignment-oriented KG embeddings for entity alignment. It is able to alleviate the negative impacts caused by the structural heterogeneity and limited seed alignments. Through two channels, MuGNN not only explicitly completes the KGs, but also pruning exclusive entities by using different relation weighting schemes: KG self-attention and cross-KG attention, showing robust graph encoding capability.
Extensive experiments on five publicly available datasets and further analysis demonstrate the effectiveness of our method.

In future, we are interested in introducing text information of entities for alignment by considering word ambiguity~\cite{cao2017modeling}; and meanwhile, through cross-KG entity proximity~\cite{cao2015name}.

\section*{Acknowledgments}

NExT++ research is supported by the National Research Foundation, Prime Minister's Office, Singapore under its IRC@SG Funding Initiative.

\bibliography{ms}
\bibliographystyle{acl_natbib}

\end{document}